\documentclass[sn-mathphys-ay]{sn-jnl}

\usepackage{graphicx}%
\usepackage{multirow}%
\usepackage{amsmath,amssymb,amsfonts}%
\usepackage{amsthm}%
\usepackage{mathrsfs}%
\usepackage[title]{appendix}%
\usepackage{xcolor}%
\usepackage{textcomp}%
\usepackage{manyfoot}%
\usepackage{booktabs}%
\usepackage{algorithm}%
\usepackage{algorithmicx}%
\usepackage{algpseudocode}%
\usepackage{listings}%
\usepackage{natbib}
\usepackage[subrefformat=parens]{subcaption}
\usepackage{bm}
\usepackage{adjustbox}

\begin{document}

\title{Machine Learning-Based Self-Localization Using Internal Sensors for Automating Bulldozers}

\author[1]{\fnm{Hikaru} \sur{Sawafuji}}

\author[2]{\fnm{Ryota} \sur{Ozaki}}

\author[2]{\fnm{Takuto} \sur{Motomura}}

\author[2]{\fnm{Toyohisa} \sur{Matsuda}}

\author[2]{\fnm{Masanori} \sur{Tojima}}

\author[1]{\fnm{Kento} \sur{Uchida}}

\author[1]{\fnm{Shinichi} \sur{Shirakawa}}

\affil[1]{\orgname{Yokohama National University}}

\affil[2]{\orgname{Komatsu Ltd}}

\abstract{Self-localization is an important technology for automating bulldozers. Conventional bulldozer self-localization systems rely on RTK-GNSS (Real Time Kinematic-Global Navigation Satellite Systems). However, RTK-GNSS signals are sometimes lost in certain mining conditions. Therefore, self-localization methods that do not depend on RTK-GNSS are required. In this paper, we propose a machine learning-based self-localization method for bulldozers. The proposed method consists of two steps: estimating local velocities using a machine learning model from internal sensors, and incorporating these estimates into an Extended Kalman Filter (EKF) for global localization. We also created a novel dataset for bulldozer odometry and conducted experiments across various driving scenarios, including slalom, excavation, and driving on slopes.
The result demonstrated that the proposed self-localization method suppressed the accumulation of position errors compared to kinematics-based methods, especially when slip occurred. Furthermore, this study showed that bulldozer-specific sensors, such as blade position sensors and hydraulic pressure sensors, contributed to improving self-localization accuracy.}


\maketitle

\section{Introduction}\label{sec1}
Bulldozers are used for construction and mining in all over the world. However, because operating bulldozers is complex and difficult, there is a shortage of skilled operators in these industries. As becoming proficient in these operations for beginners takes time, the demand for bulldozer automation has increased.
For meeting such demand, several technologies that allow beginners to operate bulldozers easily are developed.
\citet{Mononen2022} proposed semi-autonomous bulldozer blade control. Komatsu Ltd. developed and commercialized an automatic blade control for their bulldozers \citep{Kamikawa2020}.
In addition to the operational efficiency as described above, automation of bulldozers can contribute to safety of workers. Automation can make the operations unmanned, which eliminates hazards to humans. Especially, operations in mines are harsh and dangerous for human operators. \citet{Hirayama2019} proposed a path planning for unmanned bulldozers.

The automation technologies described above require self-localization. Conventional bulldozers depend on RTK-GNSS (Real Time Kinematic-Global Navigation Satellite Systems). Therefore, when the RTK-GNSS signals are lost, the automated systems cannot operate. There are several scenarios where these signals are lost. In construction, for example, the signals are more likely to be difficult to receive under bridges and near buildings. In mining, the signals are difficult to reach at the bottom of open-pit valleys. In addition, solar flares sometimes interfere with GNSS signals, which is called ionospheric scintillation.
Thus, self-localization methods that do not depend on RTK-GNSS are required.
Strapdown Inertial Navigation \citep{Titterton2004} using an IMU (Inertial Measurement Unit) is often adopted to estimate the change in the pose of mobile robots over time. However, estimating the self-pose of bulldozers only with IMUs is not easy, since running bulldozers get severe vibrations due to the rough ground and their engines. IMUs suffer from these vibrations.

For wheeled robots, the speed of movement can be estimated from the number of wheel revolutions, which is called {\it odometry}. \citet{Borenstein1996} proposed Gyrodometry which combines measurements from a gyro with odometry. However, crawlers of bulldozers often get slipped. During the slippage, the odometry estimates deviate from the actual velocities. As slips are complex phenomena, modeling them with mathematical formulae is extremely difficult.
SLAM (Simultaneous Localization and Mapping) \citep{Taheri2021} using LiDARs (Light Detection And Ranging) and/or cameras is generally another effective self-localization method for mobile robotics. However, SLAM often does not work well in environments where bulldozers work, especially in a mining environment, because there are fewer features measured by LiDARs and cameras in those environments. For example, bulldozers for mining work in environments where there are no structures and only soil all around. 

To address these issues above, in this paper, we present a novel self-localization method for bulldozers using machine learning and their internal sensors.
The proposed method utilizes the characteristics of the data from the internal sensors to address the slip of crawlers. For instance, just before slipping occurs, large static friction forces act on the crawlers. Once the crawlers start to slip, the loads on the crawlers are reduced because the static friction changes to dynamic friction. This phenomenon causes fluctuations in the measurement values of hydraulic pumps and engines. The proposed method models these features by machine learning to achieve odometry which takes slip into account.
The proposed method consists of a machine learning-based model and EKF (Extended Kalman Filter)~\citep{Ribeiro2004}.
The machine learning-based model is trained to be able to estimate the velocity of the vehicle on the local coordinate from the data of the internal sensors, such as IMU, crawler encoder, engine sensors, hydraulic sensors. Data measured from a commercialized bulldozer are used to train the model. The training data are measured when the RTK-GNSS signal is stable and the poses estimated with the RTK-GNSS and an IMU are used as ground truth.
The EKF integrates the inferred velocity on the local coordinate and the IMU data to estimate the position and posture of the bulldozer on the global coordinate.
To evaluate the proposed method, experiments using the commercialized bulldozer are conducted.
Note that, while this paper focuses on self-localization without external sensors (e.g. LiDAR, camera), the proposed method can be integrated with methods using external sensors.

The main contributions of this paper are summarized as follows:
\begin{itemize}
    \item A novel self-localization method for bulldozers using machine learning and internal sensors is presented. This is able to take slip into account by learning the characteristics of the internal sensors.
    \item A novel dataset for bulldozer odometry is created by recording data from a commercialized bulldozer. Our dataset contains data from internal sensors and components such as IMU, crawler encoder, engine, and hydraulic component. It also includes the ground truth of the pose of the bulldozer which is calculated with RTK-GNSS and an IMU.
    \item To show the effectiveness of the proposed method, experiments using the commercialized bulldozer are conducted.
\end{itemize}

\section{Related Work}\label{sec2}
Various odometry methods, such as Inertial Navigation Systems (INSs)~\citep{Titterton2004} and wheel odometry, have been proposed for self-localization in environments where GNSS is unavailable. 
INSs integrate the accelerations and angular velocities measured by an Inertial Measurement Unit (IMU) to estimate self-localization.
Wheel odometry estimates the current position and posture using the travel distance of each wheel, which is calculated by the product of wheel circumference and rotation count.
However, these dynamic models face significant challenges in practical applications due to their susceptibility to sensor noise and cumulative errors.
To address these issues of sensor noise and cumulative errors, Kalman Filters (KF)~\citep{Kalman1960} or Extended Kalman Filters (EKF)~\citep{Ribeiro2004} have been used in sensor fusion for self-localization. They provide a robust framework for estimating the state of a dynamic system from noisy measurements.
However, the effectiveness of these methods is limited to scenarios that conform to their predefined dynamic models.

Recently, machine learning-based approaches for self-localization have gained significant attention in the field of robotics and autonomous systems. 
These data-driven techniques have shown a remarkable ability to learn complex patterns and relationships from sensor data, often outperforming conventional methods. 
For example, a Multi-Layer Perceptron (MLP) neural network was used to estimate the position errors of INS during GPS outage~\citep{chiang_multisensor_2003, goodall_improving_2006}.
However, the method using MLP has the drawback of being unable to consider long-term past vehicle dynamic information. 
As a solution, \citet{fang_lstm_2020} used Long Short-Term Memory (LSTM) to consider a relationship between the present and past information.
\citet{el_sabbagh_promoting_2023} proposed a novel architecture of cascaded neural networks that considers the time dependency of velocity for low-complexity self-localization. 
In the context of wheeled vehicles, 
\citet{onyekpe_whonet_2021} proposed a novel approach for accurate positioning using the Wheel Odometry Neural Network (WhONet) to correct wheel speed errors. They evaluated the method in various challenging driving conditions, including sudden braking, sharp cornering, and wet surfaces. 
\citet{belhajem_improving_2018} used Support Vector Machines (SVM) to refine the position estimates obtained from EKF. The SVM takes wheel speed and yaw as inputs.
\citet{brossard_learning_2019} incorporated the Gaussian process regression model, which learns the residuals between the deterministic model based on wheel speed sensors or IMU and the ground truth, into an EKF.
These studies demonstrate that machine learning approaches can enable accurate self-localization by reducing accumulated position errors or correcting sensor noise.

In off-road environments, wheel slip presents an additional challenge for self-localization. 
Wheel slippage can significantly degrade localization accuracy because wheel speed measurements no longer accurately reflect the actual vehicle speed~\citep{borenstein_measurement_1996}. 
Several studies have addressed the challenge of slip in off-road environments.
\citet{yamauchi_slip-compensated_2017} classified slippage into four types based on the slippage direction and maneuver type and the proposed slip estimation methods for each type. These methods improved the localization accuracy for a tracked vehicle operating on an indoor sandy slope, outperforming conventional odometry methods. 
\citet{kim_dnn-based_2023} developed a self-localization method for wheeled mobile robots, including a DNN-based slip ratio estimator. Their approach significantly reduced localization errors compared to integration-based and EKF-based methods on a sloped grass field. 
\citet{choi_interpretable_2024} used a dual attention network for slip-aware velocity estimation from internal sensor data, such as wheel speed, acceleration, and engine torque. The proposed method was evaluated through experiments on various soil types. However, this study focused on velocity estimation and did not address self-localization. 
These machine learning-based approaches have shown promising results in capturing the complex relationships between vehicle dynamics, terrain characteristics, and slip behavior, potentially offering more adaptive and accurate solutions for slip-aware self-localization in heavy machinery like bulldozers.

While many studies have made significant advances in self-localization using internal sensors without GNSS, several crucial gaps remain, particularly in bulldozer operations. 
These gaps are significant, as bulldozer positioning involves unique challenges and requirements.
For example, many existing studies use only IMU measurements and/or wheel speed data. However, bulldozer operations often involve severe vibrations, introducing significant noise into the IMU accelerometer data. Additionally, frequent slippage occurs due to rough terrain and excavation work, leading to substantial errors in crawler speed measurements. As a result, self-localization methods that rely on an IMU and/or wheel/crawler encoders are often inaccurate for bulldozer operations. 
Moreover, bulldozers operate in unique scenarios such as excavation and slalom. These scenarios are not considered in conventional vehicle positioning research. 
These bulldozer-specific operations can significantly impact positioning accuracy, yet existing studies do not adequately address them.

\begin{figure}[tb]
    \centering
    \includegraphics[width=1\linewidth]{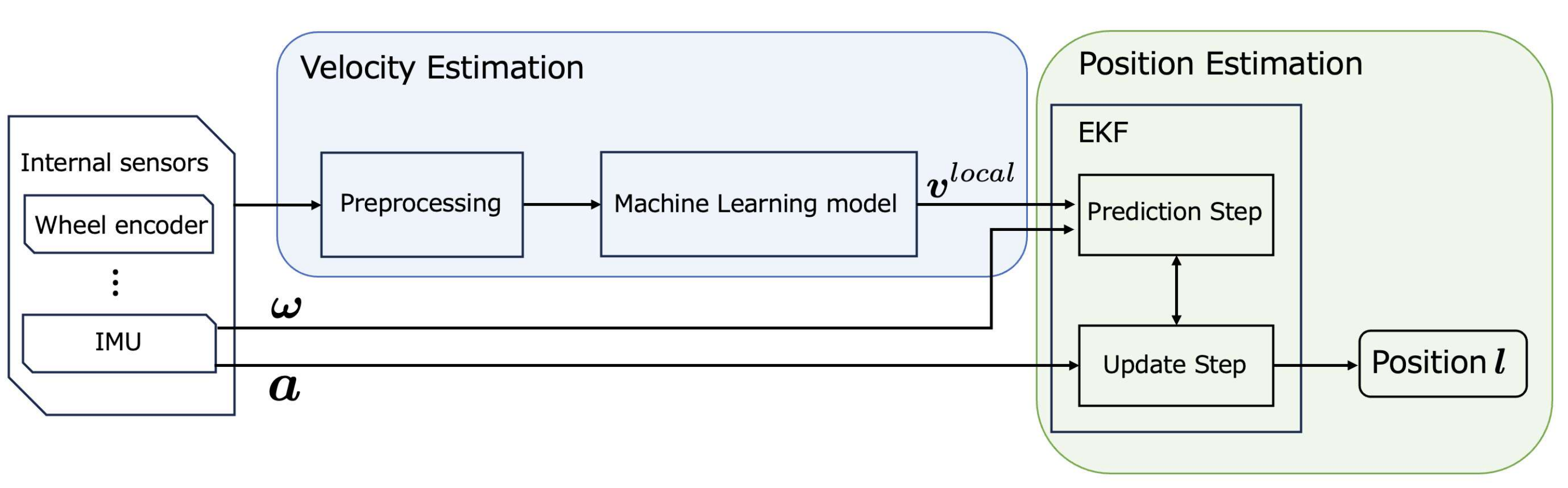}
    \caption{Overview of the proposed method}
    \label{fig:overview}
\end{figure}

\begin{figure}[tb]
    \centering
    \includegraphics[width=0.7\linewidth]{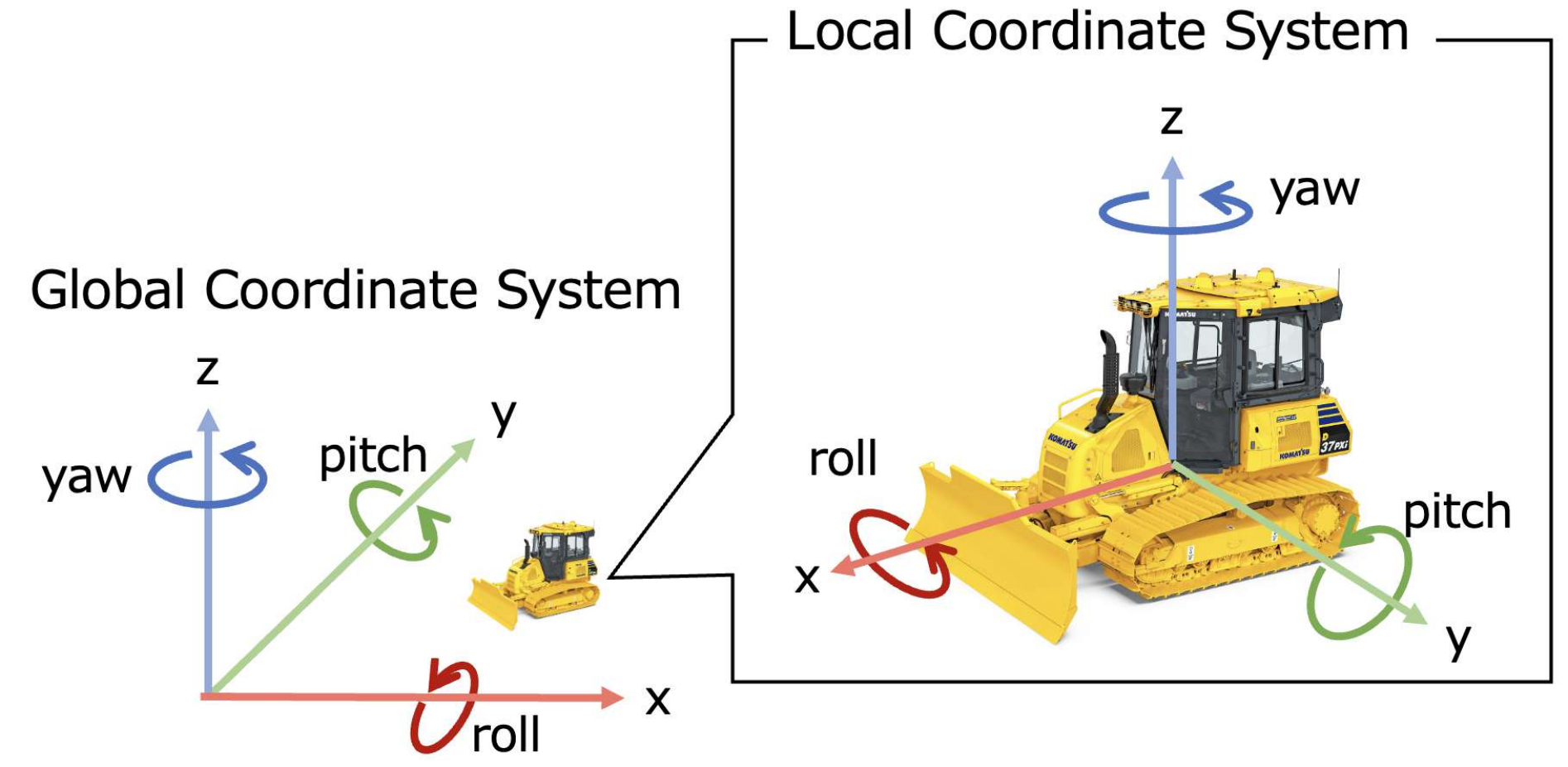}
    \caption{Relationship between global and local coordinate systems in this study.}
    \label{fig:coordinate_systems}
\end{figure}

\section{Proposed Method}\label{sec:proposed method}
We propose a machine learning-based self-localization method for bulldozers using their internal sensors. 
Figure~\ref{fig:overview} shows the overview of the proposed method. 
The proposed self-localization method for bulldozers consists of a machine learning model and an Extended Kalman Filter (EKF). 
Our method operates as follows. First, raw sensor data from various internal sensors are preprocessed.
Next, these preprocessed internal sensor data are used as input to the machine learning model to estimate the local velocity of the bulldozer. 
This estimated local velocity, along with angular velocity measurements from the IMU, serves as input to the prediction step of the EKF. The EKF then estimates the state vector containing the bulldozer's position and posture. In the update step of the EKF, acceleration data from the IMU are used to estimate the roll and pitch of the bulldozer from the gravitational acceleration, which are incorporated as observation vectors. These observation vectors are used to update the state vector estimated in the prediction step. 
Finally, the bulldozer's position in the global coordinate system is extracted from the updated state vector in the EKF's update step.
In this study, the local coordinate system is defined by a right-hand coordinate system fixed to the bulldozer, whose $x$-axis is the bulldozer's travel direction. The relationship between the global and local coordinate systems is shown in Figure~\ref{fig:coordinate_systems}.

\subsection{Velocity Estimation}\label{subsec2-1}
Our machine learning model utilizes various internal sensor data as inputs that have varying value ranges and measurement frequencies. To address these issues and adjust the data for model input, we implement two preprocessing steps:
\begin{enumerate}[1.]
\item Temporal Alignment: we perform upsampling through inserting previous measurements into missing values caused by gaps in measurement frequencies to make all input feature sequences have a consistent time interval. This process enables the model to make inferences based on the information from all sensors.
\item Standardization: following temporal alignment, we standardize each sensor's data set. This transformation adjusts the values to have a mean of zero and a standard deviation of one. Aligning the value ranges of input data may enhance the performance of neural networks~\citep{shanker_effect_1996}.
\end{enumerate}
Using these preprocessed data, the machine learning model estimates the three-axis velocity in the local coordinate system at the current time. In this study, we used three different machine learning models: the multi-layer perceptron (MLP), the Long Short Term Memory (LSTM)~\citep{hochreiter_long_1997} and the extreme gradient boosting (XGBoost)~\citep{chen_xgboost_2016}.

\subsubsection{Multi-Layer Perceptron (MLP)}
MLP is a fundamental supervised learning algorithm in neural networks. 
It is a fully connected neural network with a layered structure that consists of an input layer, one or more hidden layers, and an output layer.
The architecture of the MLP used in this study consists of 4 hidden layers with 256 units each. The input layer contains 39 units, and the output layer contains 3 units that output the estimated velocities along the $x$, $y$, and $z$ axes in the local coordinate system. The activation function for hidden layers is the ReLU function~\citep{nair_rectified_2010}. The output layer uses no activation function.

\subsubsection{Long Short Term Memory (LSTM)}
\begin{figure}[tb]
    \centering
    \includegraphics[width=\linewidth]{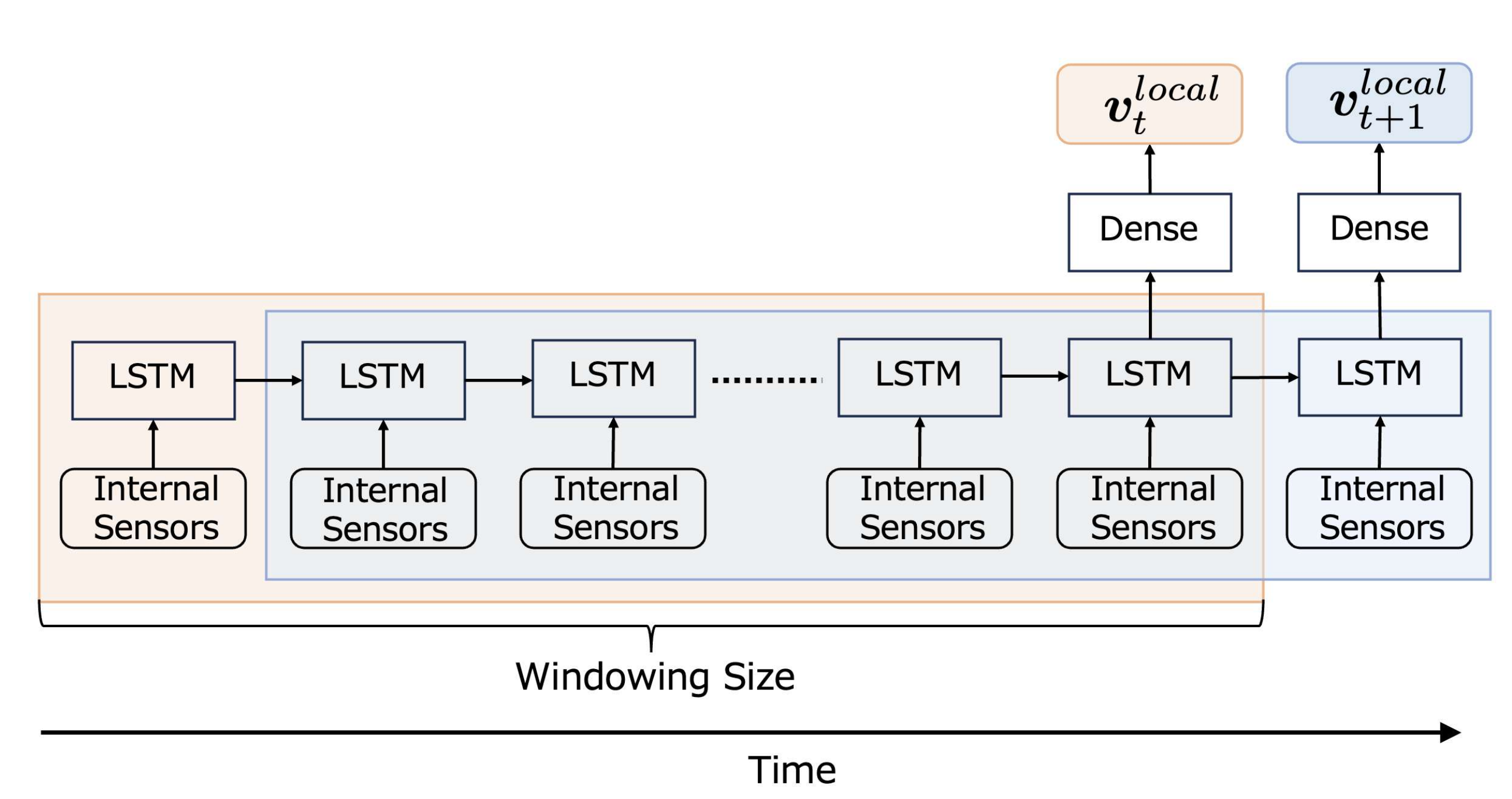}
    \caption{Windowing approach used by LSTM.}
    \label{fig:windowing approach}
\end{figure}

LSTM is a recurrent neural network that can preserve long-term memory through a gating mechanism~\citep{hochreiter_long_1997}. LSTM is known for its high performance in inference with time-series data and has been used in many related studies~\citep{fang_lstm_2020, valente_lstm_2019, liu_vehicle_2021}. In this study, we adopt the windowing approach to obtain inference values from time-series data. Figure~\ref{fig:windowing approach} demonstrates the procedure of the windowing approach. In this approach, the time-series data is split into windows of size $N$, and each window is input to LSTM. This allows LSTM to estimate the current time prediction based on past sensor data for $N$ time steps.
In this study, the windowing size is set to 40, which showed the best accuracy on the validation dataset in the preliminary experiment. The number of outputs of the LSTM layers is 256. A fully connected layer transforms the output of the LSTM layers into the estimated velocities along the $x$, $y$, and $z$ axes. The numbers of layers and hidden units of the LSTM are the same as those of MLP.

\subsubsection{Extreme Gradient Boosting (XGBoost)}
XGBoost is a typical gradient boosting method using decision trees~\citep{chen_xgboost_2016}. Boosting is an ensemble technique that constructs a strong predictive model by sequentially adding weak learners. In XGBoost, each new decision tree is trained to correct the errors of previously trained trees by fitting to their residuals. The algorithm employs gradient descent to optimize the loss function, leading to accurate prediction.
In this study, the hyperparameters are set to their default values defined in the scikit-learn package~\citep{scikit-learn}. When the prediction accuracy on the validation data does not improve over 10 iterations, the training is stopped.

\subsection{Localization}\label{subsec:Localization}
The extended Kalman filter (EKF)~\citep{Ribeiro2004} in our method estimates the current position in the global coordinate system by integrating the velocity predicted by the machine learning model with IMU data.
The EKF consists of two steps: the prediction step and the update step. In the prediction step, the state-transition model predicts the state vector and the error covariance matrix. In the update step, the state vector and the error covariance matrix are corrected using the observation model.
In the EKF, our method uses the angular velocity and acceleration measured by the IMU besides the estimated velocity by the machine learning models.

\subsubsection{Prediction Step}\label{subsubsec:Prediction Step}
In this paper, we define the position $\bm{l}_t$, posture 
$\bm{r}_t$, and state vector $\bm{s}_t$ at time $t$ as follows:
\begin{align}
\bm{l}_t &= (x_t,y_t,z_t)^\top \enspace, \quad \bm{r}_t = (\phi_t, \theta_t, \psi_t)^\top \enspace, \quad
\bm{s}_t = (\bm{l}_t,\bm{r}_t) \enspace,
\end{align}
where $x_t$, $y_t$, and $z_t$ are the $x$-, $y$-, and $z$-coordinates of the bulldozer in the global coordinate system, respectively;
$\phi_t$, $\theta_t$, and $\psi_t$ are the roll, pitch, and yaw of the bulldozer, respectively.
The changes in position $\Delta \bm{l}$ and orientation $\Delta \bm{r}$ in the global coordinate system can be calculated from the estimated velocity $\bm{v}_{t-1} \in \mathbb{R}^{3}$ and the angular velocity $\bm{\omega}_{t-1}\in \mathbb{R}^{3}$ measured by the IMU as follows:
\begin{align}
&\Delta \bm{l}_t = \mathbf{Rot}_{(\bm{r}_{t-1})}^{\mathrm{xyz}}\bm{v}_{t-1} \Delta t \enspace, \\
&\Delta \bm{r}_t = \mathbf{Rot}_{(\bm{r}_{t-1})}^{\mathrm{rpy}}\bm{\omega}_{t-1} \Delta t \enspace,
\end{align}
where $\mathbf{Rot}_{(\bm{r}_{t-1})}^{\mathrm{xyz}}\in\mathbb{R}^{3 \times 3}$  is the rotation matrix that transforms positions from the local to the global coordinate system, $\mathbf{Rot}_{(\bm{r}_{t-1})}^{\mathrm{rpy}}\in\mathbb{R}^{3 \times 3}$ is the matrix that transforms angular velocities from the local coordinate system to orientation changes in the global coordinate system, and $\Delta t$ is the sampling interval of the internal sensors.
The rotation matrices $\mathbf{Rot}_{(\bm{r}_{t})}^{\mathrm{xyz}}$ and $\mathbf{Rot}_{(\bm{r}_{t})}^{\mathrm{rpy}}$ are explicitly defined as follows~\citep{beard_quadrotor_2008}:
\begin{gather}
\mathbf{Rot}_{(\bm{r}_{t})}^{\mathrm{xyz}} =
\begin{pmatrix}
C_{\theta_{t}}C_{\psi_{t}} & S_{\phi_{t}}S_{\theta_{t}}C_{\psi_{t}} - C_{\phi_{t}}S_{\psi_{t}} & C_{\phi_{t}}S_{\theta_{t}}C_{\psi_{t}} + S_{\phi_{t}}S_{\psi_{t}} \\
C_{\theta_{t}}S_{\psi_{t}} & S_{\phi_{t}}S_{\theta_{t}}S_{\psi_{t}} + C_{\phi_{t}}C_{\psi_{t}} & C_{\phi_{t}}S_{\theta_{t}}S_{\psi_{t}} - S_{\phi_{t}}C_{\psi_{t}} \\
-S_{\theta_{t}} & S_{\phi_{t}}C_{\theta_{t}} & C_{\phi_{t}}C_{\theta_{t}}
\end{pmatrix},\\
\mathbf{Rot}_{(\bm{r}_{t})}^{\mathrm{rpy}} = 
\begin{pmatrix}
1 & S_{\phi_{t}}T_{\theta_{t}} & C_{\phi_{t}}T_{\theta_{t}} \\
0 & C_{\phi_{t}} & -S_{\phi_{t}} \\
0 & S_{\phi_{t}}/C_{\theta_{t}} & C_{\phi_{t}}/C_{\theta_{t}}
\end{pmatrix},
\end{gather}
where \(C_{\xi}\), \(S_{\xi}\), and \(T_{\xi}\) are short for \(\cos\xi\), \(\sin\xi\), and \(\tan\xi\) with $\xi \in \{ \phi_t, \theta_t, \psi_t \}$, respectively. 
The position $\bm{l}_{t}$ and posture $\bm{r}_{t}$ are calculated as follows:
\begin{align}
\bm{l}_t = \bm{l}_{t-1} + \Delta \bm{l}_t \enspace, \qquad \bm{r}_t =  \bm{r}_{t-1} + \Delta \bm{r}_t \enspace.
\end{align}
Defining the state transition model $f$, the state vector $\bar{\bm{s}}_t$ at the current time step $t$ can be predicted as follows:
\begin{align}
&\bar{\bm{s}}_t = f(\bm{s}_{t-1},\bm{u}_{t-1}) = \bm{s}_{t-1} + 
    \begin{pmatrix}
        \mathbf{Rot}_{(\mathbf{r}_{t-1})}^{\mathrm{xyz}}\bm{v}_{t-1} \Delta t\\
        \mathbf{Rot}_{(\mathbf{r}_{t-1})}^{\mathrm{rpy}}\bm{\omega}_{t-1} \Delta t 
    \end{pmatrix}\enspace,
\end{align}
where $\bm{s}_t$ is the state vector after the previous update step, and $\bm{u}_t$ is the control vector given by
\begin{align}
&\bm{u}_{t} = 
    \begin{pmatrix}
     \bm{v}_{t} \Delta t\\
     \bm{\omega}_{t} \Delta t 
    \end{pmatrix}\enspace.
\end{align}
The error covariance matrix $\bar{\mathbf{P}}_t\in \mathbb{R}^{6 \times 6}$ at the current time step is predicted as follows:
\begin{align}
    &\bar{\mathbf{P}}_t = \mathbf{J}_{f_{t-1}}\mathbf{P}_{t-1}\mathbf{J}_{f_{t-1}}^\top
                        + \mathbf{Q}_{t-1} 
    \qquad \text{where} \quad \mathbf{J}_{f_{t-1}} = \left.\frac{\partial f}{\partial \bm{s}}\right|_{\bm{s}_{t-1}, \bm{u}_{t-1}} \enspace .
\end{align}
Here, $\mathbf{J}_{f_{t-1}}\in \mathbb{R}^{6 \times 6}$ is the Jacobian matrix of the state transition function, $\mathbf{P}_{t-1}\in \mathbb{R}^{6 \times 6}$ is the error covariance matrix at the previous update step, and $\mathbf{Q}_{t-1}\in \mathbb{R}^{6 \times 6}$ is the process noise covariance matrix.

\subsubsection{Update Step}\label{subsubsec:Update Step}
Using the acceleration measured by the IMU, the predicted state vector $\bar{\bm{s}}_{t}$ and the predicted error covariance matrix $\bar{\mathbf{P}}_{t}$ are updated.
Assuming that the bulldozer is at rest or in constant velocity linear motion, the roll and pitch of the bulldozer can be calculated from the acceleration $\mathbf{a}=(a_x, a_y, a_z)^\top$ as follows:
\begin{align}
    \mathbf{z} &= 
    \begin{pmatrix}
        \phi_{\mathrm{obs}}, \theta_{\mathrm{obs}}
    \end{pmatrix}^\top
    =
    \begin{pmatrix}
        \tan^{-1} \left( {\frac{a_y}{a_z}} \right), \tan^{-1} \left( {\frac{a_x}{\sqrt{a_y^2+a_z^2}}} \right)
    \end{pmatrix}^\top.
\end{align}
We treat $\mathbf{z}$ as the observation vector.
The observation model $h$ transforms the state vector into the observation vector as follows:
\begin{align}
    h(\bar{\bm{s}}_{t})=
    \begin{pmatrix}
        0,0,0,1,0,0\\
        0,0,0,0,1,0
    \end{pmatrix}
    \bar{\bm{s}}_t \enspace .
\end{align}
Using the observation vector $\mathbf{z}$ and the observation model $h$, the predicted state vector $\bar{\bm{s}}_{t}$ and predicted error covariance matrix $\bar{\mathbf{P}}_{t}$ are updated by 
\begin{align}
    &{\bm{s}}_t = \bar{\bm{s}}_t + 
                        \mathbf{K}_t(\mathbf{z}_t-h(\bar{\bm{s}}_t)) \enspace ,\\
    &{\mathbf{P}}_t = (\mathbf{I}_6-\mathbf{K}_t\mathbf{J}_{h_t})\bar{\mathbf{P}}_t \enspace ,
\end{align} 
where $\mathbf{K}_t\in \mathbb{R}^{6 \times 2}$ is the Kalman gain, $\mathbf{I}_6 \in \mathbb{R}^{6 \times 6}$ is the identity matrix, and $\mathbf{J}_{h_t}\in \mathbb{R}^{2 \times 6}$ is the Jacobian matrix of the observation model. The Kalman gain and the Jacobian matrix of the observation model are given by the following equations:
\begin{align}
    \mathbf{K}_t &= \bar{\mathbf{P}}_t\mathbf{J}_{h_{t}}^{\top}(\mathbf{J}_{h_{t}}\bar{\mathbf{P}}_t\mathbf{J}_{h_{t}}^{\top}+\mathbf{R}_t)^{-1} \quad \text{and} \quad
    \mathbf{J}_{h_{t}} =\left.\frac{\partial h}{\partial \bm{s}}\right|_{\bm{s}_{t}} \enspace, 
\end{align}
where $\mathbf{R}_t\in \mathbb{R}^{2 \times 2}$ is the measurement noise covariance matrix.
Our method outputs the state vector ${\bm{s}}_t$ and the error covariance matrix ${\mathbf{P}}_t$ obtained in the observation step.

\begin{figure}[tb]
    \centering
    \includegraphics[width=0.3\linewidth]{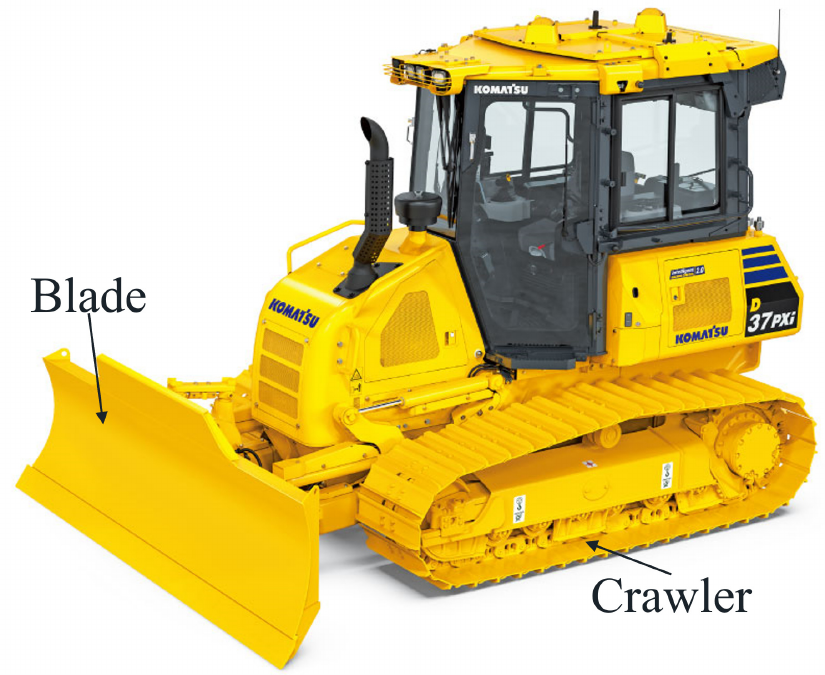}
    \caption{Komatsu D37PXI-24. We used this bulldozer to collect data.}\label{fig: d37}
\end{figure}

\section{Dataset}\label{sec:dataset}
Data measured from a bulldozer are collected in order to train and evaluate the proposed method on them.
In this paper, a Komatsu bulldozer, D37PXI-24 \citep{Fukasawa2004} in Figure \ref{fig: d37}, is used to collect a dataset. This bulldozer adopts Hydro Static Transmission (HST) \citep{Hayashi2008} that rotates the crawlers with hydraulic pumps. This bulldozer has an IMU and two GNSS antennas to localize itself.
The dataset contains the features listed in Table \ref{tab: list_of_data_contained_in_dataset}. The pose of the vehicle is estimated using GNSS and IMU when the GNSS signals are stably received. This estimated pose data is used as the target variable and is not input into the proposed model. Each data listed in Table \ref{tab: list_of_data_contained_in_dataset} is measured at 10 Hz or 100 Hz. For example, the IMU data are measured at 100 Hz, while the crawler velocities are measured at 10 Hz.

\begin{table}[tb]
    \caption{List of features contained in dataset}\label{tab: list_of_data_contained_in_dataset}
    \begin{tabular*}{\linewidth}{@{\extracolsep\fill}lp{4cm}cc}
        \toprule%
        Description & Content & Frequency & Category\footnotemark[1] \\
        & & [Hz] & \\
        \midrule
        Time difference from previous step & [s] & 100 & IC \\
        Vehicle pose\footnotemark[2] & \(x\) [m], \(y\) [m], \(z\) [m] & 100 & IC \\
        Acceleration measured by IMU & \(x\) [\(\mathrm{m/s^2}\)], \(y\) [\(\mathrm{m/s^2}\)], \(z\) [\(\mathrm{m/s^2}\)] & 100 & IC \\
        Angular velocity measured by IMU & \(x\) [deg/s], \(y\) [deg/s], \(z\) [deg/s] & 100 & IC \\
        Posture measured by IMU & roll [deg], pitch [deg], yaw [deg] & 100 & IC \\
        Crawler velocity & right [m/h], left [m/h] & 10 & IC \\
        Speed gear status & 0: 1st, 1: 2nd, 2: 3rd & 10 & Ve \\
        FNR gear status & 0: forward, 1: neutral, 2: reverse & 100 & Ve \\
        Steering state & 0: no turning, 1: turning right, 2: turning left & 10 & Ve \\
        Steering lever stroke & [\%] & 100 & Ve \\
        Blade lift lever stroke & [\%] & 100 & Bu \\
        Blade tilt lever stroke & [\%] & 100 & Bu \\
        Blade state & 0: still, 1: lifting, 2: lowering, 3: right tilting, 4: left tilting, 5: right angling, 6: left angling & 10 & Bu \\
        Command current to blade & lift [mA], tilt [mA], angle [mA] & 100 & Bu \\
        Blade right cutting edge position\footnotemark[3] & \(x\) [mm], \(y\) [mm], \(z\) [mm] & 100 & Bu \\
        Blade left cutting edge position\footnotemark[3] & \(x\) [mm], \(y\) [mm], \(z\) [mm] & 100 & Bu \\
        Blade lift angle & [deg] & 10 & Bu \\
        Engine speed & [rpm] & 100 & Ve \\
        Engine torque & [Nm] & 100 & Ve \\
        HST pump pressure \citep{Hayashi2008} & right front [MPa], left front [MPa], right rear [MPa], left rear [MPa] & 10 & Bu \\
        Blade pump pressure & [MPa] & 10 & Bu \\
        Relief level of blade pump pressure & 0: less than 36 MPa, 1: 36 MPa or more & 10 & Bu \\
        Traction force\footnotemark[4] & [W] & 10 & Ve \\
        \botrule
    \end{tabular*}
    \footnotetext[1]{In this paper, the features are categorized into three groups: IMU and Crawler velocity only (IC), sensors typically available in vehicles (Ve), and bulldozer-specific sensors (Bu). These categories are used in the ablation study in Section \ref{subsec5-3}}
    \footnotetext[2]{The pose of the vehicle is estimated using GNSS and IMU. It is represented in the global coordinate system. This data is used as the target variable and is not input into the model.}
    \footnotetext[3]{The position of the blade cutting edge is represented in the local coordinate system.}
    \footnotetext[4]{The traction force is calculated based on the pressure, speed, and capacity of the hydraulic pumps.}
\end{table}

\begin{figure}[tb]
    \begin{tabular}{ccccc}
        \begin{minipage}[t]{0.19\linewidth}
            \centering
            \includegraphics[keepaspectratio, width=1.0\linewidth]{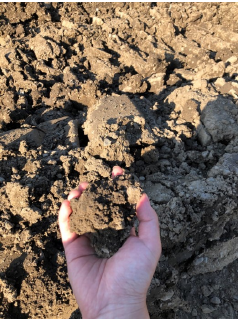}
            \vspace*{0.01cm}
            \subcaption{Clay 1}
        \end{minipage}
        &
        \begin{minipage}[t]{0.19\linewidth}
            \centering
            \includegraphics[keepaspectratio, width=1.0\linewidth]{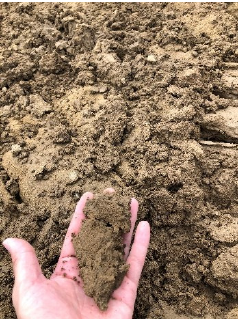}
            \vspace*{0.01cm}
            \subcaption{Clay 2}
        \end{minipage}
        &
        \begin{minipage}[t]{0.19\linewidth}
            \centering
            \includegraphics[keepaspectratio, width=1.0\linewidth]{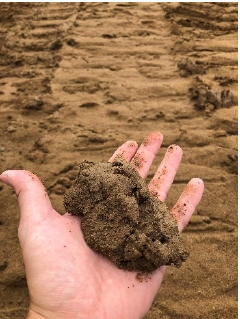}
            \vspace*{0.01cm}
            \subcaption{Sandy soil}
        \end{minipage}
        &
        \begin{minipage}[t]{0.19\linewidth}
            \centering
            \includegraphics[keepaspectratio, width=1.0\linewidth]{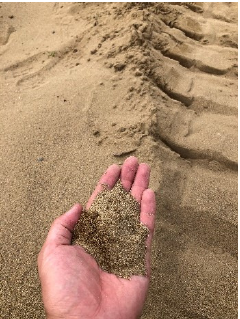}
            \vspace*{0.01cm}
            \subcaption{Sand}
        \end{minipage}
        &
        \begin{minipage}[t]{0.19\linewidth}
            \centering
            \includegraphics[keepaspectratio, width=1.0\linewidth]{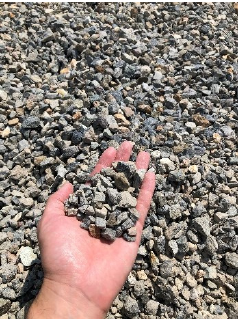}
            \vspace*{0.01cm}
            \subcaption{Gravel}
        \end{minipage} 
    \end{tabular}
    \caption{Fields for collecting data.}
    \label{fig: fields}
\end{figure}

\begin{figure}[tb]
    \begin{tabular}{cccc}
        \begin{minipage}[t]{0.25\linewidth}
            \centering
            \includegraphics[keepaspectratio, width=0.95\linewidth]{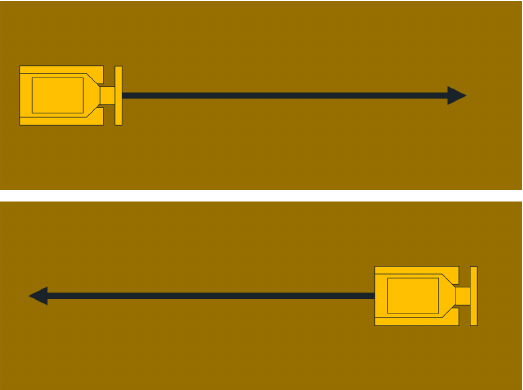}
            \subcaption{Straight}
            \label{fig:straight}
        \end{minipage}
        &
        \begin{minipage}[t]{0.25\linewidth}
            \centering
            \includegraphics[keepaspectratio, width=0.95\linewidth]{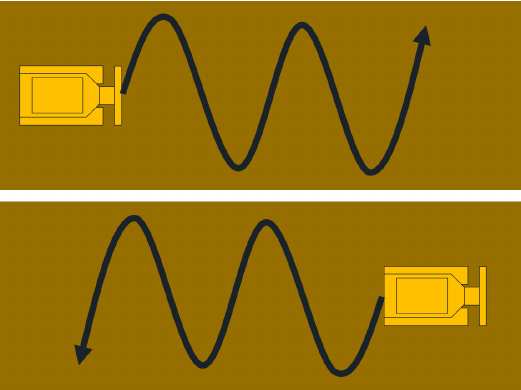}
            \subcaption{Low-frequency slalom}
            \label{fig:low-frequency slalom}
        \end{minipage}
        &
        \begin{minipage}[t]{0.25\linewidth}
            \centering
            \includegraphics[keepaspectratio, width=0.95\linewidth]{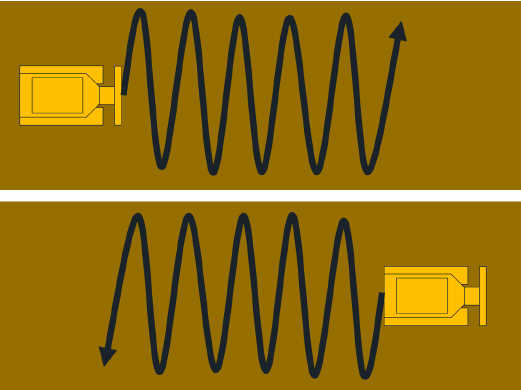}
            \subcaption{High-frequency slalom}
            \label{fig:high-frequency slalom}
        \end{minipage}
        &
        \begin{minipage}[t]{0.25\linewidth}
            \centering
            \includegraphics[keepaspectratio, width=0.95\linewidth]{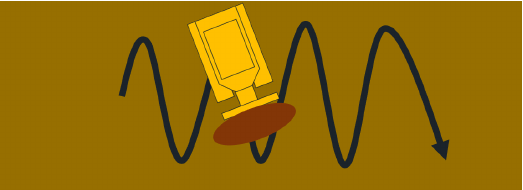}
            \subcaption{Slalom while carrying dirt}
            \label{fig:slalom while carrying dirt}
        \end{minipage}
        \\
        \begin{minipage}[t]{0.25\linewidth}
            \centering
            \includegraphics[keepaspectratio, width=0.95\linewidth]{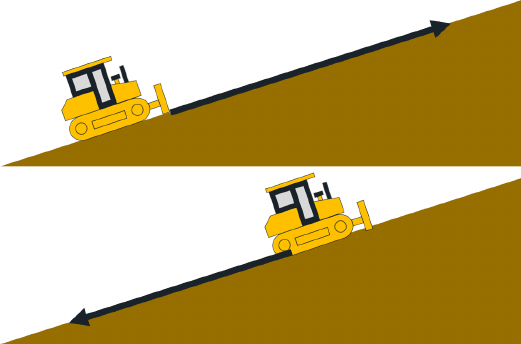}
            \subcaption{Climbing slope}
            \label{fig:climbing slope}
        \end{minipage}
        &
        \begin{minipage}[t]{0.25\linewidth}
            \centering
            \includegraphics[keepaspectratio, width=0.95\linewidth]{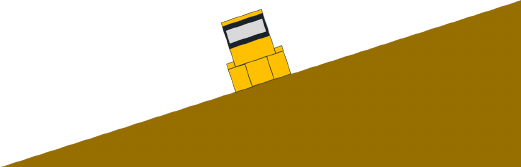}
            \subcaption{Crossing slope}
            \label{fig:crossing slope}
        \end{minipage}
        &
        \begin{minipage}[t]{0.25\linewidth}
            \centering
            \includegraphics[keepaspectratio, width=0.95\linewidth]{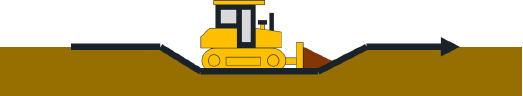}
            \subcaption{Excavation}
            \label{fig:excavation}
        \end{minipage}
        &
        \begin{minipage}[t]{0.24\linewidth}
            \centering
            \includegraphics[keepaspectratio, width=0.95\linewidth]{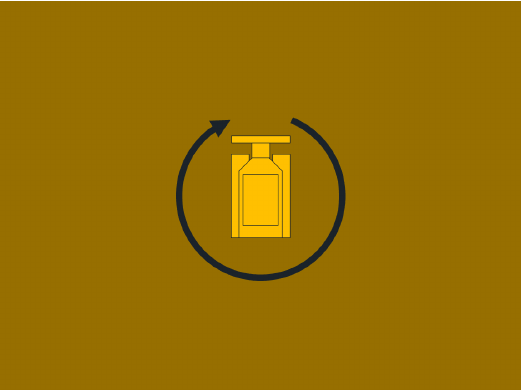}
            \subcaption{Turn}
            \label{fig:turn}
        \end{minipage}
        \\
        \begin{minipage}[t]{0.25\linewidth}
            \centering
            \includegraphics[keepaspectratio, width=0.95\linewidth]{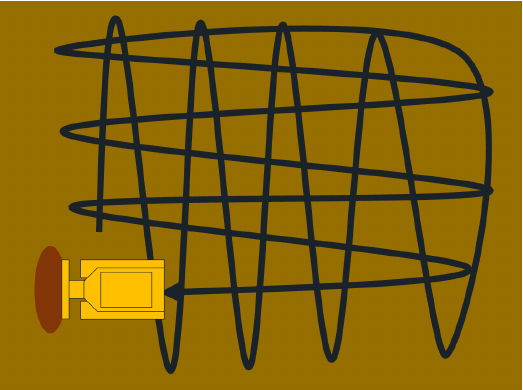}
            \subcaption{Grading}
            \label{fig:grading}
        \end{minipage}
        &
        \begin{minipage}[t]{0.25\linewidth}
            \centering
            \includegraphics[keepaspectratio, width=0.95\linewidth]{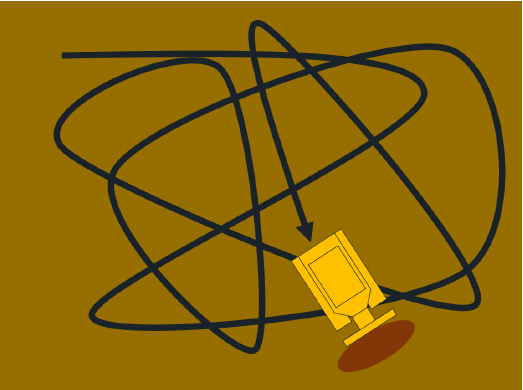}
            \subcaption{Random}
            \label{fig:random}
        \end{minipage}
    \end{tabular}
    \caption{Driving scenarios in our dataset.}
    \label{fig: drive_patterns}
\end{figure}

The data collection took place in five areas. These areas have different soil types as shown in Figure~\ref{fig: fields}. The sandy soil has properties intermediate between sand and clay.
The bulldozer was operated in 10 different driving scenarios as shown in Figure~\ref{fig: drive_patterns}. In total, 462 episodes were recorded, which is approximately 13 hours of data. The duration of each episode is approximately 30 to 200 seconds.

\section{Experiments}\label{sec5}
In this section, we evaluated the proposed method described in Section~\ref{sec:proposed method} with the bulldozer-driving data described in Section~\ref{sec:dataset}. 
We conducted two types of evaluation experiments: one for velocity estimation and another for localization. 
In the velocity estimation evaluation, we compared three machine learning models, MLP, LSTM, and XGBoost, with the kinematic calculation from the crawler encoder. 
In the localization evaluation, we compared the performance of conventional rule-based methods with the proposed learning-based methods. We conducted experiments across various driving scenarios typical for bulldozers.

\subsection{Experimental Setup}\label{sec:exp setup}
The training data consists of 442 sequences. Each sequence contains between 3,000 and 20,000 time steps of data. We used about 47,000,000 data for training the model. The test data contains 10 sequences that were chosen from each bulldozer-driving scenario described in Section~\ref{sec:dataset}. The validation data was selected in the same manner as the test data. 

The MLP and LSTM were optimized using the Adam optimizer \citep{kingma_adam_2015} with an initial learning rate of 0.001. We employed the Mean Squared Error (MSE) as the loss function to train the models over 100 epochs with a mini-batch size of 2048. We validated the models using the validation dataset every five epochs, and the best-performing model in terms of the loss function value was used for evaluation. XGBoost was trained using the default settings provided by scikit-learn 1.3.2~\citep{scikit-learn}.

The process noise covariance matrix $\mathbf{Q}$ is set as a diagonal matrix with values corresponding to the uncertainties in position and orientation, specifically $\mathbf{Q}=\mathrm{diag}(0.1 \times (\Delta t)^2,0.1 \times (\Delta t)^2,0.1 \times (\Delta t)^2,0.01 \times (\Delta t)^2,0.01 \times (\Delta t)^2,0.01 \times (\Delta t)^2)$, where the first three elements represent the noise for the $x$, $y$, and $z$ positions, and the latter three represent the noise for the roll, pitch, and yaw.
Similarly, the measurement noise covariance matrix $\mathbf{R}$ is configured as a diagonal matrix with $\mathbf{R}=\mathrm{diag}(0.01,0.01)$, reflecting the uncertainties in the measurement data.

We conducted our experiments on NVIDIA Jetson AGX Orin 32GB platform with the following specifications: 1792-core NVIDIA Ampere architecture GPU with 56 Tensor Cores and an 8-core Arm Cortex-A78AE v8.2 64-bit CPU at 2.2GHz.

\subsection{Evaluation of Velocity Estimation}\label{subsec5-2}
To evaluate the performance of the velocity estimation using machine learning models, we compared the velocities estimated by the machine learning models with the raw velocity measured by the crawler encoder.
The crawler encoder measures the track rotation, and the velocity is kinematically calculated assuming that the bulldozer is running on the flat ground without slippage.
Note that the crawler encoder can only measure $x$-velocities and assumes $y$- and $z$-velocities are zero.
As the evaluation metric, we used the Root Mean Square Error (RMSE). We calculated the RMSE for each direction ($x$-, $y$-, and $z$-axes) between the estimated and the ground truth velocities. The ground truth velocities are calculated based on the position information obtained from the RTK-GNSS.

\begin{figure}[tb]
    \centering
    \begin{minipage}[c]{\linewidth}
        \centering
        \includegraphics[width=1\linewidth]{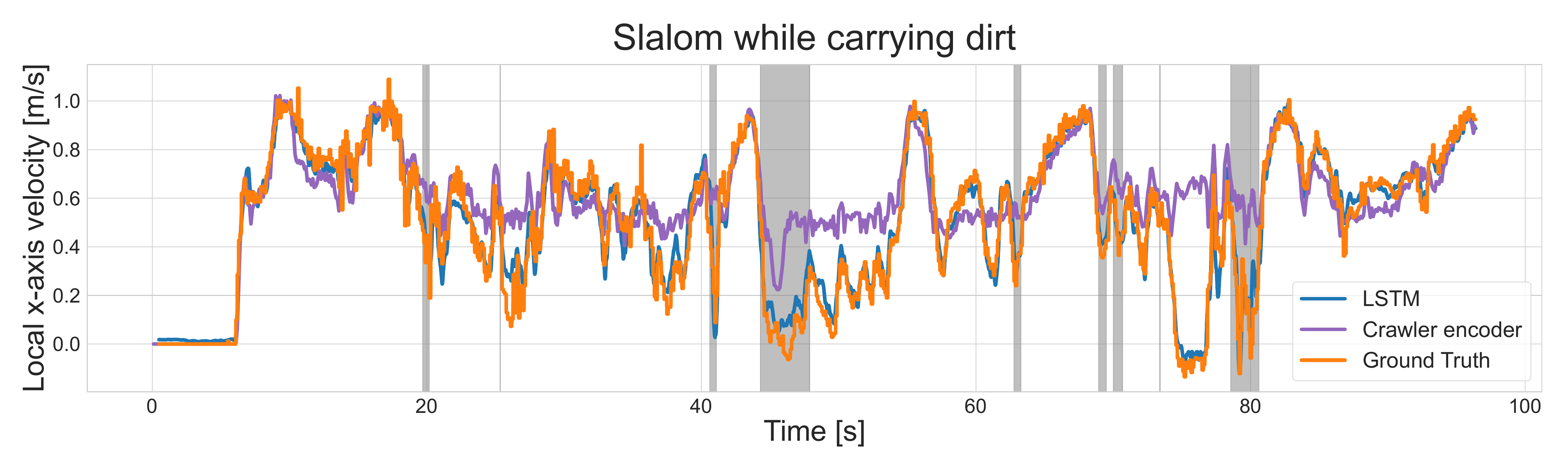}
        \label{fig:local_xvel_vs_time(short_slalom_while_excavating)}
    \end{minipage}
    \begin{minipage}[c]{\linewidth}
        \centering
        \includegraphics[width=1\linewidth]{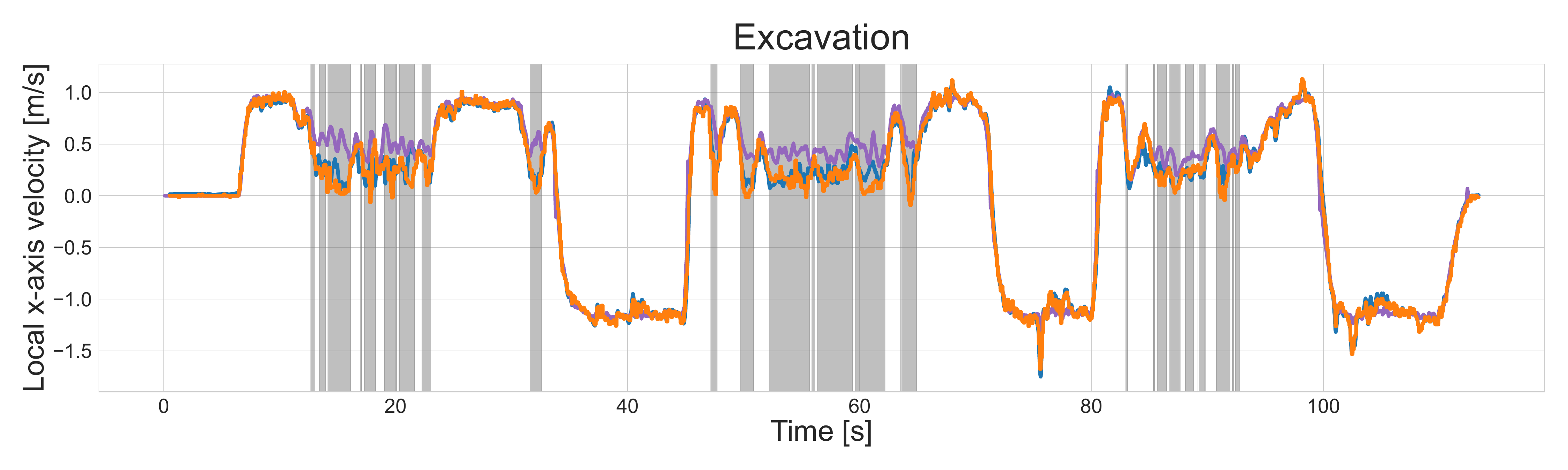}
        \label{fig:local_xvel_vs_time(excavation)}
    \end{minipage}
    \caption{Plot of local velocities along $x$-axis during the slalom while carrying dirt (see Figure \ref{fig:slalom while carrying dirt}) and excavation (see Figure \ref{fig:excavation}). The sections shaded in gray indicate periods when the bulldozer is slipping. }
    \label{fig:local_xvel_vs_time}
\end{figure}

Table~\ref{table:velocityEstimationEval} shows the results. It can be seen that the velocities estimated by the machine learning models had smaller errors than the raw velocities measured by the crawler encoder. 
Moreover, among the LSTM, MLP, and XGBoost, the LSTM was the most accurate model in estimating the local velocities. This result implies that the past vehicle dynamic information, which was only addressed by the LSTM, is important for estimating the local velocities.
Figure \ref{fig:local_xvel_vs_time} shows the plot of local velocities along $x$-axis during slalom while carrying dirt and excavation. The sections shaded in gray represent periods when the bulldozer was slipping.
During these periods, the velocity measured by the crawler encoder was higher than the true velocity, indicating that, while the crawler tracks were rotating, the bulldozer itself was not moving forward. 
On the other hand, the velocity estimated by LSTM closely matched the true velocity even during the slipping periods.
Note that this slip flag estimated with RTK-GNSS was not used in either the proposed methods or the conventional methods. 
It indicates that the proposed machine learning models such as LSTM learned to adjust velocity during slipping periods.

\begin{table}[tb]
    \captionsetup{width=\linewidth}
    \caption{Comparison of raw velocity measured by crawler encoder and velocity estimated by machine learning models. We report the averages and standard deviations calculated from five independent trials.}
    \label{table:velocityEstimationEval}
    \centering
    \begin{tabular}{lccc}
        \toprule
        & \multicolumn{3}{@{}c@{}}{RMSE[m/s]}\\
        \cmidrule{2-4}
        Method & $x$-axis & $y$-axis & $z$-axis \\
        \midrule
        Crawler encoder & $0.2469$ & $0.0687$ & $0.0219$ \\
        LSTM(ours) & $\bm{0.0590}\pm0.0014$ & $\bm{0.0426}\pm0.0002$ & $\bm{0.0154}\pm0.0002$ \\
        MLP(ours) & $0.0764\pm0.0013$ & $0.0509\pm0.0009$ & $0.0180\pm0.0002$ \\
        XGBoost(ours) & $0.0873$ & $0.0543$ & $0.0189$ \\
        \botrule
    \end{tabular}
\end{table}

\subsection{Evaluation of Localization}\label{subsec5-3}
We used a simple crawler odometry method and a kinematics-based EKF as comparative methods. These methods did not use a machine learning model. 
In the crawler odometry method, we calculated the forward velocity $v^{\mathrm{local}}_{\mathrm{x}}$ and the angular velocity around the $z$-axis $\omega_{\mathrm{z}}$ from the velocities measured by the left and right crawler encoders as follows:
\begin{align}
    \label{eq:local_x_vel}
    v^{\mathrm{local}}_{\mathrm{x}} = \frac{1}{2}(v^{\mathrm{local}}_{\mathrm{xr}}+v^{\mathrm{local}}_{\mathrm{xl}}) \qquad 
    \omega_{\mathrm{z}} = \frac{1}{T}(v^{\mathrm{local}}_{\mathrm{xr}}-v^{\mathrm{local}}_{\mathrm{xl}}),
\end{align}
where $v^{\mathrm{local}}_{\mathrm{xl}}$ and $v^{\mathrm{local}}_{\mathrm{xr}}$ are the velocities measured by the left and right crawler encoders, respectively, and \(T \) represents the distance between the crawler tracks (i.e.~tread), which was set to 2.77 [m] in this study. This crawler odometry assumes that the velocity in the $y$-axis and $z$-axis and the angular velocity around the $x$-axis and $y$-axis in the local coordinate system were zero. The movement amount at each time step is calculated from the calculated velocity and angular velocity, and the position is estimated by accumulating these movement amounts from the initial position.

The kinematics-based EKF uses the velocity measured by the crawler encoders as the local $x$-velocities for the prediction step within the EKF. This kinematics-based EKF assumes that the velocities in the $y$- and $z$-directions in the local coordinate system were zero. The other components and input values within the EKF were the same as those in the proposed method, which are described in Section \ref{subsec:Localization}.

As the evaluation metrics, we used Average Displacement Error (ADE):
\begin{align}
\text{ADE} = \frac{1}{N} \sum_{i=1}^{N} \| \hat{\bm{l}}_i - \bm{l}_i \|_2 \enspace.
\end{align}
Here, $N$ represents the number of predicted points, $\hat{\bm{l}}_i$ denotes the predicted position, and $\bm{l}_i$ indicates the true position.

\begin{table*}[tb]
    \caption{Comparison of the proposed self-localization with the kinematic models in various driving scenarios. The result with the best average value is represented in bold.}
    \label{table:PositionEstimationEval}
    \centering
    \begin{adjustbox}{width=\linewidth}
    \begin{tabular}{lccccc}
        \toprule%
        & \multicolumn{5}{@{}c@{}}{Average Displacement Error (ADE) [m]}\\
        \cmidrule{2-6}
        Driving Scenario & Crawler Odometry & Kinematics-based EKF & LSTM w. EKF (ours) & MLP w. EKF (ours) & XGBoost w. EKF (ours)\\
        \midrule
        Straight & $1.62$ & $0.44$ & $0.33 \pm 0.07$ & $0.35 \pm 0.05$ & $\bm{0.32}$ \\
        Low-frequency slalom & $20.79$ & $1.58$ & $0.75 \pm 0.11$ & $\bm{0.64} \pm 0.07$ & $0.81$ \\
        High-frequency slalom & $12.22$ & $15.87$ & $\bm{0.97} \pm 0.22$ & $1.72 \pm 0.29$ & $2.08$ \\
        Slalom while carrying dirt & $1.49$ & $2.73$ & $\bm{0.66} \pm 0.10$ & $0.77 \pm 0.20$ & $0.74$ \\
        Climbing slope & $2.82$ & $0.49$ & $0.30 \pm 0.04$ & $0.27 \pm 0.03$ & $\bm{0.22}$ \\
        Crossing slope & $3.78$ & $0.93$ & $0.27 \pm 0.08$ & $0.30 \pm 0.07$ & $\bm{0.21}$ \\
        Excavation & $5.08$ & $5.07$ & $\bm{0.33} \pm 0.12$ & $1.03 \pm 0.53$ & $2.00$ \\
        Turn & $0.20$ & $0.25$ & $0.10 \pm 0.08$ & $\bm{0.06} \pm 0.02$ & $0.08$ \\
        Grading & $3.85$ & $\bm{0.24}$ & $0.33 \pm 0.06$ & $0.36 \pm 0.07$ & $0.30$ \\
        Random & $29.14$ & $7.08$ & $0.93 \pm 0.19$ & $0.93 \pm 0.19$ & $\bm{0.92}$ \\
        \midrule
        Average & $8.10$ & $3.47$ & $\bm{0.50} \pm 0.05$ & $0.64 \pm 0.06$ & $0.77$ \\
        \bottomrule
    \end{tabular}
    \end{adjustbox}
\end{table*}

\begin{figure}[tb]
    \centering
    \begin{minipage}[c]{1.0\linewidth}
        \centering
        \includegraphics[width=1\linewidth]{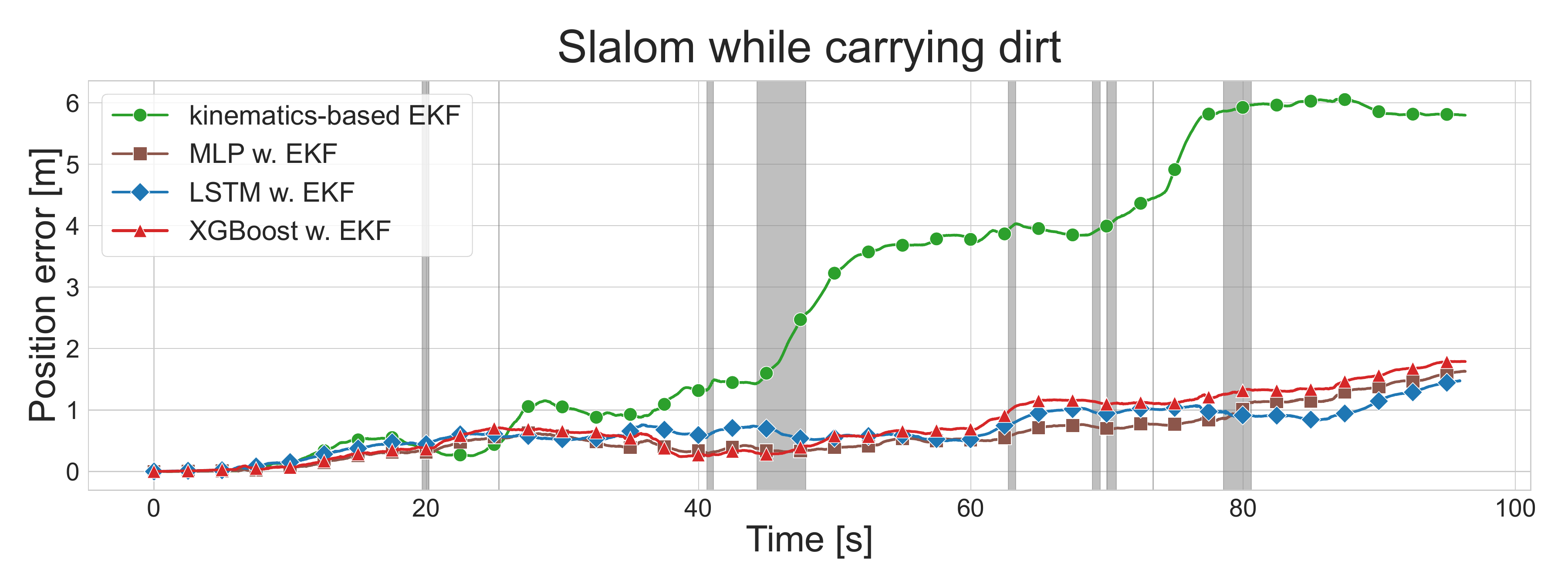}
        \label{fig:pos_error_vs_time(slalom)}
    \end{minipage}
    \begin{minipage}[c]{1.0\linewidth}
        \centering
        \includegraphics[width=1\linewidth]{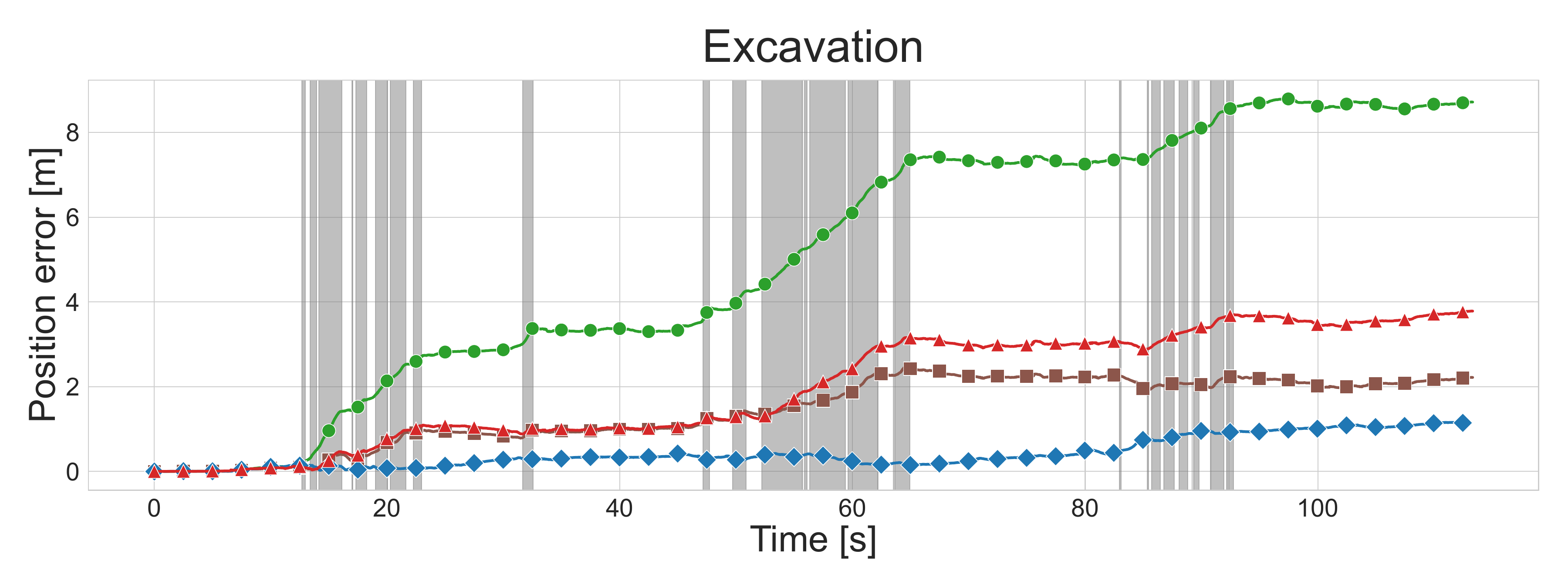}
        \label{fig:pos_error_vs_time(excavation)}
    \end{minipage}
    \caption{The position error of the trajectories in the slalom while excavating (top) and excavation (bottom). The sections shaded in gray indicate periods when the bulldozer is slipping.}
    \label{fig:pos_error_vs_time}
\end{figure}

\begin{figure}
    \centering
    \begin{minipage}[c]{1.0\linewidth}
        \centering
        \includegraphics[width=1\linewidth]{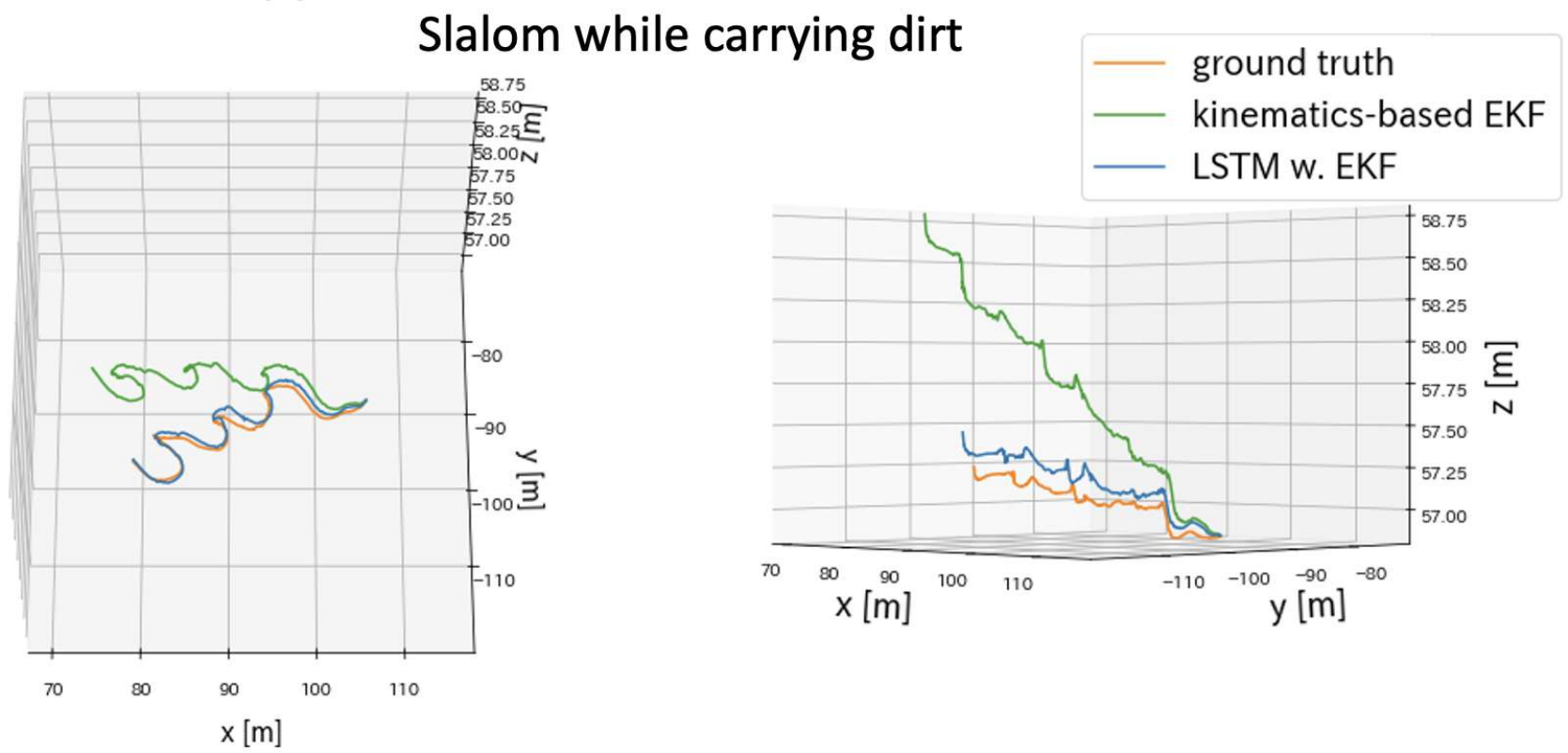}
        \label{fig:trajectory(slalom)}
    \end{minipage}
    \begin{minipage}[c]{1.0\linewidth}
        \centering
        \includegraphics[width=1\linewidth]{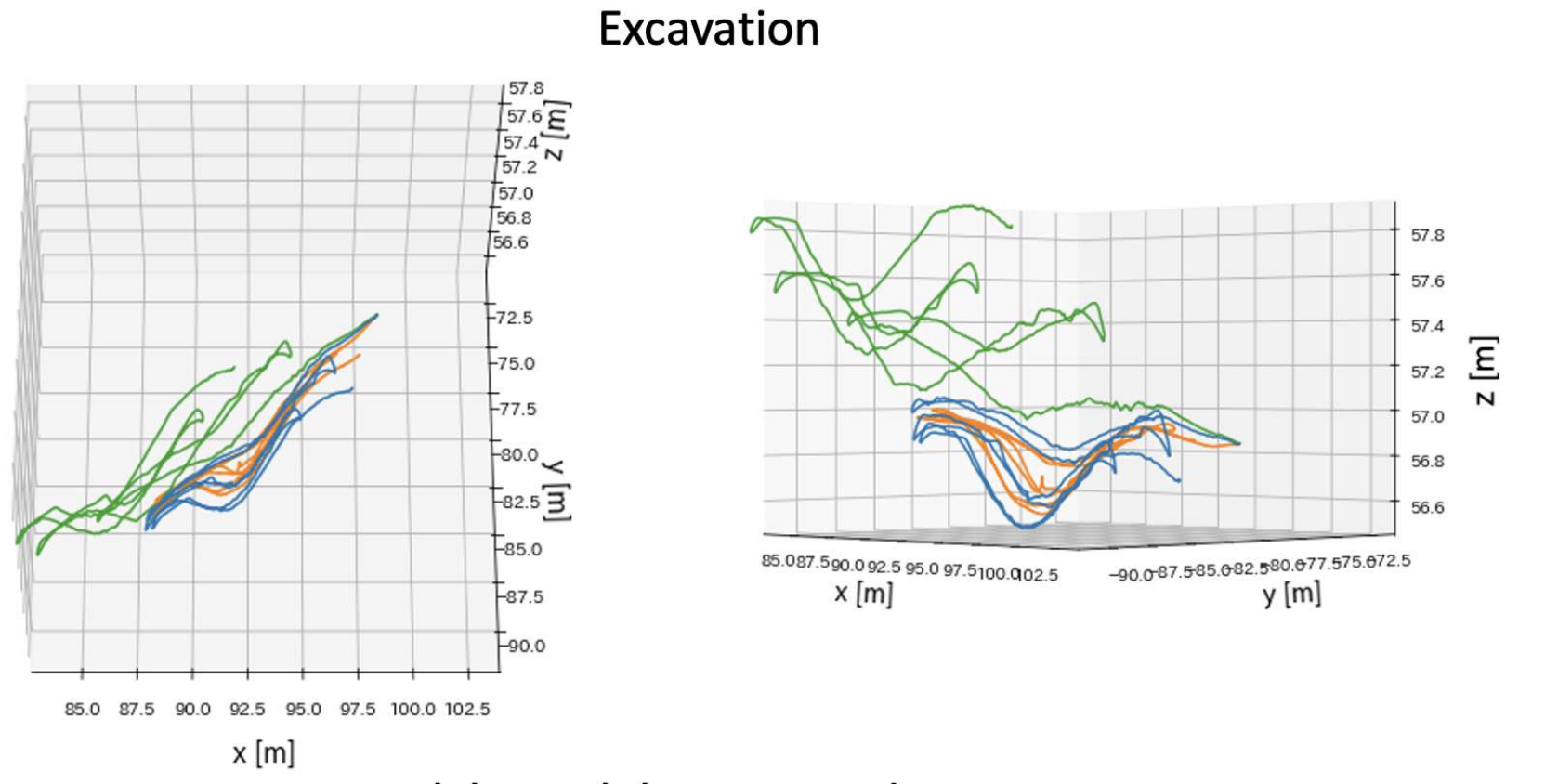}
        \label{fig:trajectory(excavation)}
    \end{minipage}
    \caption{The trajectories in the slalom while carrying dirt (top) and excavation (bottom). }
    \label{fig:trajectory}
\end{figure}

Table \ref{table:PositionEstimationEval} presents the localization results across various driving scenarios.
The proposed methods estimated the positions more accurately than the comparison methods in all driving scenarios except for grading. 
It was observed that the best machine learning model differs depending on the driving scenarios. While the XGBoost~w.~EKF demonstrated superior performance in slope-related driving scenarios such as climbing and crossing slopes (see Figures \ref{fig:climbing slope} and \ref{fig:crossing slope}), the LSTM~w.~EKF achieved the lowest errors in driving scenarios where slippage occurs frequently, such as high-frequency slalom, slalom while carrying dirt, and excavation (see Figures \ref{fig:high-frequency slalom}, \ref{fig:slalom while carrying dirt}, and \ref{fig:excavation}). 
Figure~\ref{fig:pos_error_vs_time} shows the position error of the trajectories estimated by the kinematics-based EKF (green line), MLP w. EKF (brown line), LSTM w. EKF (blue line) and XGBoost w. EKF (red line).  By using machine learning models for velocity estimation, the position error at the final time step was effectively reduced. When the bulldozer was slipping (painted gray in Figure~\ref{fig:pos_error_vs_time}), the position error of the kinematics-based EKF tended to increase significantly. In contrast, the proposed method, which uses a machine learning model for velocity estimation, suppresses the expansion of errors even when the bulldozer was slipping.
Figure~\ref{fig:trajectory} presents the trajectory-estimation results of the kinematics-based EKF (green line) and LSTM w. EKF (blue line) and the ground truth trajectory (orange line) in the slalom while carrying dirt and excavation. The LSTM w. EKF estimated a trajectory closer to ground truth than the kinematics-based EKF.

Additionally, to assess the feasibility of real-time self-localization, we evaluated the computation time for each method. The computation time was measured using data collected from a random driving scenario lasting 343.39 seconds. We conducted the verification using the same computer described in Section \ref{sec:exp setup}. Table \ref{table:computation time} presents the results.
Although the use of machine learning models significantly increased the computation time compared to conventional methods like crawler odometry and kinematics-based EKF, all proposed methods demonstrated capability for real-time application. Even LSTM w. EKF, which had the longest computation time, took only 207.99 seconds to process the entire 343.39 seconds of driving data. Given that the data were collected at a frequency of 100 Hz, we can calculate the average processing time per frame: 207.99 s / 34339 frames $\fallingdotseq$ 0.0061 s/frame. This demonstrates that the LSTM w. EKF method still processes each frame faster than the 0.01 s interval between frames in real-time data collection.
These results confirm that all tested methods, including our machine learning-enhanced approaches, are suitable for real-time self-localization applications in bulldozer operations.

\begin{table}
    \centering
    \captionsetup{width=\linewidth}
    \caption{Comparison of computation time for a random driving scenario. The data for this scenario was measured over a driving time of 343.39 seconds.}
    \label{table:computation time}
    \begin{tabular}{lcccccc}
        \toprule
         & Computation Time [s] \\
        \midrule
        Crawler Odometry & $7.30$ \\
        kinematics-based EKF & $12.40$ \\
        MLP w. EKF (ours) & $99.89$ \\
        LSTM w. EKF (ours) & $207.99$ \\
        XGBoost w. EKF (ours) & $72.37$ \\
        \botrule
    \end{tabular}
\end{table}

\subsection{Ablation Study for Evaluation of Features}
To analyze the importance of bulldozer-specific sensor data in estimating the position of a bulldozer across different driving scenarios, we conducted experiments using varying combinations of input sensor data. We categorized the sensor inputs into three groups: IMU measurements and Crawler velocity only (IC), sensors typically available in vehicles (Ve), and bulldozer-specific sensors (Bu). Table \ref{tab: list_of_data_contained_in_dataset} provides more details.
Several studies on vehicle self-localization use IC-sensors as the minimum required sensor inputs. Ve-sensors contain sensors for engine speed, engine torque, and speed gear status. Bu-sensors include blade position and pump hydraulic pressure.
We then increased the types of sensor inputs to the machine learning model, i.e., IC, IC+Ve, and IC+Ve+Bu, to observe how these additions affected localization accuracy. 

Table \ref{table:result of different input feature combinations} presents the results of this analysis, showing the ADE for each sensor combination across different driving scenarios.
The results indicate that incorporating additional sensor data generally improves the localization accuracy. The mean ADE across all scenarios decreased from 0.73 $\pm$ 0.09 [m] with IC only, to 0.63 $\pm$ 0.02 [m] with IC+Ve, and further to 0.50 $\pm$ 0.05 [m] with IC+Ve+Bu. The result also suggests that bulldozer-specific sensors contribute to improving localization accuracy, particularly in scenarios such as high-frequency slalom, excavation, and random (see Figures \ref{fig:high-frequency slalom}, \ref{fig:excavation}, and \ref{fig:random}).
The pump hydraulic pressure, included among the bulldozer-specific sensors, can capture changes in the friction force between the ground and the crawler during slippage. We consider that incorporating pump hydraulic pressure data into the machine learning model can lead to accurate self-localization, especially during the frequent slippages that occur in scenarios such as high-frequency slalom, excavation, and random.

\begin{table}
    \centering
    \captionsetup{width=\linewidth}
    \caption{Comparison of self-localization accuracy of LSTM w. EKF with different input feature combinations. We report the averages and standard deviations calculated from three independent trials. The result with the best average value is represented in bold.}
    \label{table:result of different input feature combinations}
    \begin{tabular}{lccc}
        \toprule
        & \multicolumn{3}{@{}c@{}}{Average Displacement Error (ADE) [m]}\\
        \cmidrule{2-4}
        Driving Scenario & IC & IC+Ve & IC+Ve+Bu\\
        \midrule
        Straight & $0.33 \pm 0.06$ & $\bm{0.30} \pm 0.08$ & $0.33 \pm 0.07$ \\
        Low-frequency slalom & $\bm{0.56} \pm 0.10$ & $0.63 \pm 0.13$ & $0.75 \pm 0.11$ \\
        High-frequency slalom & $1.52 \pm 0.18$ & $1.39 \pm 0.29$ & $\bm{0.97} \pm 0.22$ \\
        Slalom while carrying dirt & $0.71 \pm 0.14$ & $\bm{0.63} \pm 0.11$ & $0.66 \pm 0.10$ \\
        Climbing slope & $\bm{0.22} \pm 0.05$ & $0.36 \pm 0.14$ & $0.30 \pm 0.04$ \\
        Crossing slope & $0.50 \pm 0.05$ & $0.42 \pm 0.10$ & $\bm{0.27} \pm 0.08$ \\
        Excavation & $1.86 \pm 0.30$ & $0.77 \pm 0.39$ & $\bm{0.33} \pm 0.12$ \\
        Turn & $\bm{0.07} \pm 0.03$ & $0.18 \pm 0.15$ & $0.10 \pm 0.08$ \\
        Grading & $\bm{0.32} \pm 0.07$ & $0.38 \pm 0.07$ & $0.33 \pm 0.06$ \\
        Random & $1.23 \pm 0.29$ & $1.21 \pm 0.22$ & $\bm{0.93} \pm 0.19$ \\
        \midrule
        Average & $0.73 \pm 0.09$ & $0.63 \pm 0.02$ & $\bm{0.50} \pm 0.05$ \\
    \botrule
    \end{tabular}
\end{table}

\section{Conclusion}\label{sec6}
In this study, we proposed a machine learning approach for the self-localization of bulldozers using internal sensors. Various internal sensor values (e.g., IMU, crawler encoder, engine, hydraulic component) were input to a machine learning model to estimate the velocities in the local coordinate system. These estimated velocities, along with the angular velocities and accelerations measured by the IMU, are then used in an EKF to estimate the position. We evaluated the performance of the proposed method using data from a wide range of bulldozer tasks, including slalom and excavation. Evaluation experiments demonstrated that using machine learning for velocity estimation can suppress the increase in velocity estimation error even while the bulldozer is slipping, resulting in accurate localization. Additionally, it was found that bulldozer-specific sensors contribute to bulldozer localization, particularly in driving scenarios where slippage occurs frequently.

In this study, we did not employ complex models to ensure real-time performance. Future research will focus on progressively increasing the complexity of the models, aiming to enhance localization accuracy without compromising real-time performance.
In addition, we will analyze how the inclusion of bulldozer-specific sensors influences bulldozer localization. Understanding the impact of these features will allow us to refine our models and develop more robust and precise localization methods.

\bibliography{sn-bibliography}

\end{document}